\documentclass[conference]{IEEEtran}
\IEEEoverridecommandlockouts
\usepackage{algorithm}
\usepackage{algorithmic}
\usepackage{graphicx}
\usepackage{cite}
\usepackage{amsmath,amssymb,amsfonts}
\usepackage{array}
\usepackage{mathtools}
\newtheorem{definition}{Definition}
\usepackage{textcomp}
\usepackage{xcolor}
\def\BibTeX{{\rm B\kern-.05em{\sc i\kern-.025em b}\kern-.08em
    T\kern-.1667em\lower.7ex\hbox{E}\kern-.125emX}}
\def\mathbi#1{\textbf{\em #1}}
\begin{document}

\title{Temporal Self-Attention Network for Medical Concept Embedding
% *\\
}

\author{
\IEEEauthorblockN{
    Xueping Peng\IEEEauthorrefmark{1},
    Guodong Long\IEEEauthorrefmark{1},
    Tao Shen\IEEEauthorrefmark{1},
    Sen Wang\IEEEauthorrefmark{2},
    Jing Jiang\IEEEauthorrefmark{1},
    Michael Blumenstein\IEEEauthorrefmark{1}
}
\IEEEauthorblockA{
    \IEEEauthorrefmark{1} Centre for Artificial Intelligence, FEIT, University of Technology Sydney, Australia \\ 
    \IEEEauthorrefmark{2} School of Information Technology and Electrical Engineering, The University of Queensland, Australia \\
    Email: \{xueping.peng, guodong.long\}@uts.edu.au, Tao.Shen@student.uts.edu.au, \\ sen.wang@uq.edu.au, \{jing.jiang, Michael.Blumenstein\}@uts.edu.au}
}

\maketitle

\begin{abstract}
In longitudinal electronic health records (EHRs), the event records of a patient are distributed over a long period of time and the temporal relations between the events reflect sufficient domain knowledge to benefit prediction tasks such as the rate of inpatient mortality. Medical concept embedding as a feature extraction method that transforms a set of medical concepts with a specific time stamp into a vector, which will be fed into a supervised learning algorithm. The quality of the embedding significantly determines the learning performance over the medical data. In this paper, we propose a medical concept embedding method based on applying a self-attention mechanism to represent each medical concept. We propose a novel attention mechanism which captures the contextual information and temporal relationships between medical concepts. A light-weight neural net, ``Temporal Self-Attention Network (TeSAN)'', is then proposed to learn medical concept embedding based solely on the proposed attention mechanism. To test the effectiveness of our proposed methods, we have conducted clustering and prediction tasks on two public EHRs datasets comparing TeSAN against five state-of-the-art embedding methods. The experimental results demonstrate that the proposed TeSAN model is superior to all the compared methods. To the best of our knowledge, this work is the first to exploit temporal self-attentive relations between medical events. 
\end{abstract}

\section{Introduction}\label{intro}

\begin{figure}
\centering
\includegraphics[width=0.4\textwidth]{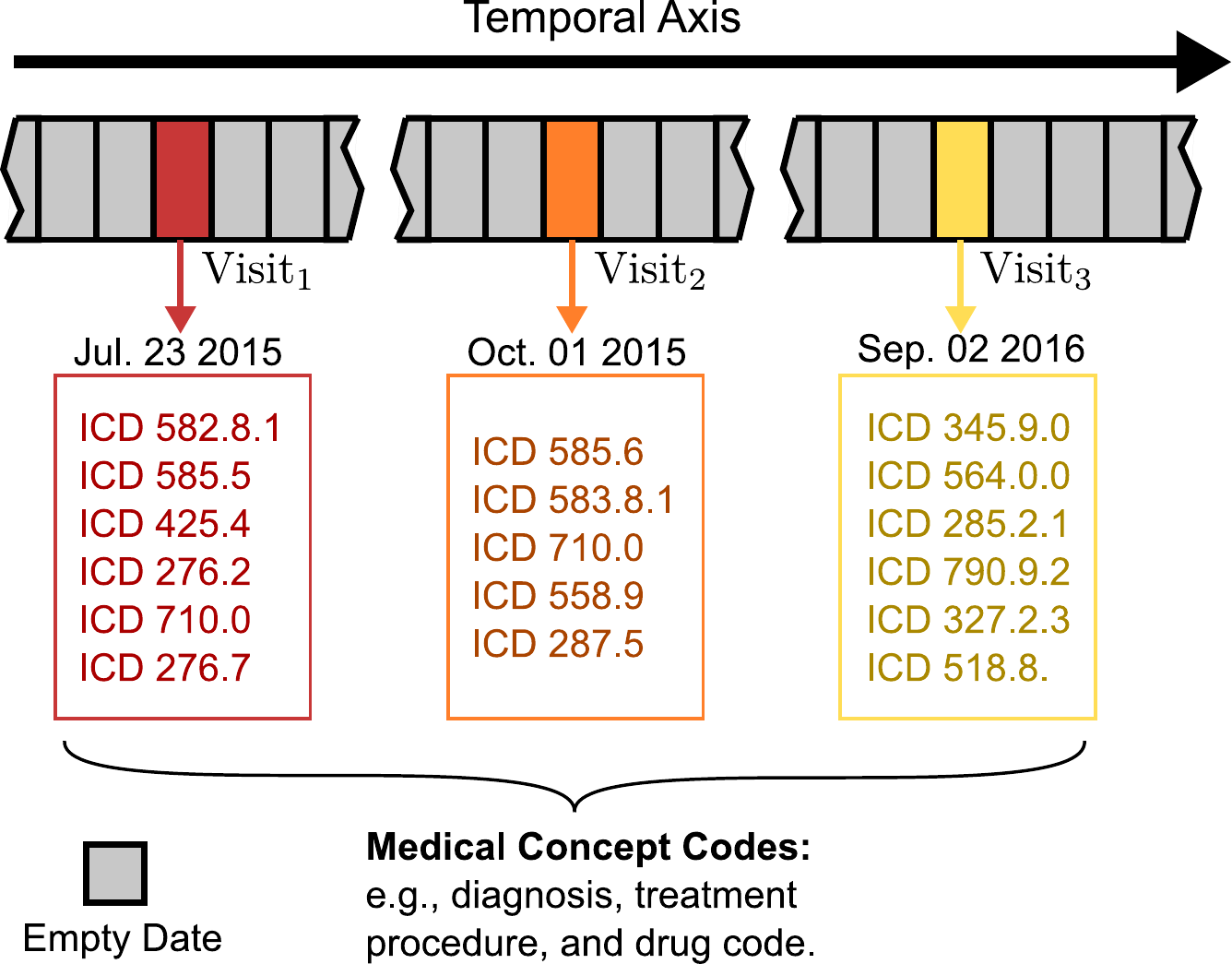}
\caption{An example segment of one patient's healthcare journey} \label{ehr-fig1}
\end{figure}

A healthcare information system (HIS) stores huge volumes of Electronic Health Records (EHRs) that contain detailed visit information about patients over a period of time ~\cite{Shickel_2018}. The EHRs data is a multi-layer structure composed of three layers: patient, visit, and medical concept. For instance, an anonymous patient in Fig. \ref{ehr-fig1} makes three visits in different days. The first and third visits recorded a diagnosis of six health conditions (denoted by diagnosis codes, e.g., ICD 585.5) while the second visit reports five disorders. A patient's healthcare journey (referred to hereafter as ``patient journey''), can thus be represented by a sequence of visits occurring at different time-stamps. To standardize the healthcare procedure, medical concepts (referred to in this paper as ``diseases'') in each visit record are converted to an item in a standard coding system (e.g., International Classification of Diseases or ICD\footnote{http://www.icd9data.com}). A medical coding system is often developed according to disease ontology and represented by a hierarchical structure, which is practical for human understanding and maintenance. This tree-based coding system includes basic medical taxonomy knowledge which can be embedded into a unified learning framework to achieve better classification performance and interpretability. In the light of this idea, a medical concept embedding method for subsequent learning tasks is highly desirable. 

Intuitively, one-hot encoding of medical concepts simply generates a binary vector that is high-dimensional and sparse. An alternative solution, inspired by Natural Language Processing (NLP), is to use word embedding approaches to learn a low-dimensional dense representation of medical concepts ~\cite{Bengio_2003,Collobert_2008,Mikolov_2013,Pennington_2014}. This method has been used in various AI-based  healthcare applications~\cite{Minarro_2014,De_Vine_2014,Tran_2015,Choi_AMIA_2016,Choi_Bahadori_2017,zhang2018integrative} to improve performance. However, there are two major limitations. First, even though there is a similar multi-layer structure in a textual corpus (document, sentence, and word) compared to EHRs, intrinsic differences are still evident. For instance, two consecutive sentences in one document only have a sequential relationship, while two arbitrary visits in one patient journey may be separated by different time intervals, which is an important factor in longitudinal studies. In other words, the period of time between two visits, which have been largely disregarded in the existing works on medical concept embedding, can be modeled as auxiliary information fed into the supervised algorithms.
% \textcolor{red}{In other words, the period of time between two visits, which have been largely disregarded in the existing works on medical concept embedding, can be modeled as auxiliary information fed into the supervised algorithms.}
Also, a sentence may include repeated words, whereas the medical concept in the visit is unique. Hence, the existing NLP models, such as word embedding and sentence embedding, cannot be directly applied to encode the medical concepts without information loss. 

Second, tree-based embedding methods cannot reflect the complex relationships between each unit of the medical concepts because of the hierarchical representation. In an EHRs dataset, there are many complicated sequential co-occurrence relationships between medical concepts that contain much richer information than tree-based taxonomy. For example, chronic kidney disease (585.5) and end stage renal disease (585.6) are separately encoded  in ICD9. Both medical records and pathology support the fact that these two diseases are temporally correlated. In other words, chronic kidney disease often eventually leads to end stage renal disease. Therefore, an encoding method that considers temporal information between medical concepts over time significantly benefits prediction tasks in healthcare analytics. References~\cite{choi2018mime,Choi_2016} proposed a multi-level representation learning that simultaneously incorporates visits and medical concepts using the sequential order of visits and the co-occurrence of medical concepts. Reference~\cite{MCE_Cai_2018} proposed a CBOW-based medical concept embedding method enhanced by an attention mechanism to capture the temporal relation between visits. In particular, the temporal sequence of patient visits has been split into many time units (e.g., week, month, and year) so that the attention mechanism can capture the sequential information as well as the time-aware information. However, a fixed size of time units is impractical because a different diagnosis or treatment might have different awareness of time. Moreover, large time units may cause information loss because it puts several visits into one time unit. Furthermore, the time intervals between visits are used as quantitative scalars to segment time units~\cite{MCE_Cai_2018} and quantify the attribute relevance~\cite{Rajkomar_Google_2018}. Although improved performance is achieved by using the quantitative scalars on healthcare analysis tasks, these aspects of time-aware methods are arbitrary and unsmooth.
%for example, diagnoses prediction~\cite{Choi_Bahadori_2017,Choi_2016,Qiao_2018}, predicting inpatient mortality~\cite{Rajkomar_Google_2018} and length of stay after admission~\cite{Rajkomar_Google_2018}.

% References~\cite{Rajkomar_Google_2018} and \cite{MCE_Cai_2018} proposed time-aware attention mechanisms to capture the relationships between medical concepts in a patient's visits. The time intervals between visits are used as quantitative scalars to segment time units~\cite{MCE_Cai_2018} and quantify the attribute relevance~\cite{Rajkomar_Google_2018}. Although improved performance is achieved by using the quantitative scalars on healthcare analysis tasks, these aspects of time-aware methods are arbitrary and unsmooth. In this study, we transform the time intervals as vectors whose dimension is the same as that of the embedded medical concept. Therefore, it is more expressive and smoother than a time scalar for capturing temporal relationships in medical concepts.

To overcome the aforementioned limitations and consider the representation of time intervals, we propose a novel attention mechanism, called ``Temporal Self-Attention (TeSA)", for temporal context fusion. In particular, we first transform the time intervals as vectors whose dimension is the same as that of the embedded medical concept. Therefore, it is more expressive and smoother than a time scalar for capturing temporal relationships in medical concepts. Then a proposed self-attention mechanism is utilized to capture the contextual information and temporal interval between medical concepts in context, and to apply a feature fusion gate to combine the attentive outputs with the original inputs to produce the final context-aware representations of all the medical concepts. A light-weight neural network based on TeSA, called ``Temporal Self-Attention Network (TeSAN)'', is also developed. TeSAN uses attention pooling to compress the output of TeSA into a vector representation. In experiments, we compare TeSAN with the state-of-the-art methods in both unsupervised and supervised learning tasks, which are clustering (i.e. nearest neighbour search) and mortality tasks, respectively. TeSAN achieves the highest normalized mutual information (NMI) and Precision at 1 (P@1) on two public medical data sets, MIMIC III and CMS, and obtains the best performance of PR-AUC and ROC-AUC for the mortality prediction task on MIMIC III data. 

The remainders of this paper are organized as follows. Section ~\ref{Relat} reviews related studies. In Section ~\ref{Preliminary}, we briefly discuss some preliminary, and details about our model are presented in Section ~\ref{Method}. In Section ~\ref{Experim}, we demonstrate the experimental results conducted on two public datasets. Lastly, we conclude our study in Section ~\ref{Conc}.%and outline our future work

\section{Related Work}\label{Relat}
\subsection{Word Embedding}
Although word embedding was first introduced by Rumelhart et al.~\cite{Rumelhart_1986} in 1986, distributed representation learning of words with neural networks has only become a hot research topic since 2003~\cite{Bengio_2003,Mikolov_2013,Mikolov_2013_b,Collobert_2008,Pennington_2014,Bojanowski_2017}. CBOW and the Skip-gram model~\cite{Mikolov_2013,Mikolov_2013_b} are among two of the model families that were introduced to compute continuous vector representations of words from very large datasets. Each is based on the assumption that the order of words or a word's context do not influence the projection of the target word. However, some scholars have recently discovered that sequence and context do matter. For example, Melamud et al.~\cite{Melamud_2016} explored the impact of context with the Skip-gram model, finding that weighting for context improves performance with extrinsic tasks. Similarly, Liu et al.~\cite{Liu_2017} showed that conditioning a target word on a subset of contexts improves both the quality of the embedding and the predictions. Ling et al.~\cite{Ling_2015} extended CBOW by incorporating an attention model that considers contextual words and their positions relative to the predicted word, which results in better representations. Each of these advancements has proven effective in the field of NLP but, as discussed in Section~\ref{intro}, the differences between documents and patient journeys mean these embedding models cannot be directly applied to medical concepts in EHRs without information loss or reduced performance.

\subsection{Medical Concept Embedding}
Borrowing ideas from word representation models~\cite{Mikolov_2013,Mikolov_2013_b,jha2018interpretable}, researchers in the healthcare domain have recently explored the possibility of creating representations of medical concepts. Much of this research has focused on the Skip-gram model. For example, Minarro-Gimnez et al.~\cite{Minarro_2014} directly applied Skip-gram to learn representations of medical text, and Vine et al.~\cite{De_Vine_2014} did the same for UMLS medical concepts. Choi et al.~\cite{Choi_AMIA_2016} went a step further and used the Skip-gram model to learn medical concept embeddings from different data sources, including medical journals, medical claims, and clinical narratives. In other work~\cite{Choi_2016}, Choi et al. developed the Med2Vec model based on Skip-gram to learn concept-level and visit-level representations simultaneously. The shortcoming of all these models is that they view EHRs as documents in the NLP sense, which means that temporal information is ignored. 

Attention mechanisms are a more recent introduction in healthcare analytics~\cite{Bahdanau_2014,lee2018diagnosis}. Choi et al.~\cite{Choi_Bahadori_2017} proposed a graph-based attention model that learns representations of medical concepts from medical ontologies. Rajkomar et al.~\cite{Rajkomar_Google_2018} applied an attention-based time-aware neural network model~\cite{liu2019prototype} to predict patient outcomes, and Cai et al.~\cite{MCE_Cai_2018} proposed MCE (Medical Concept Embedding) as a way to integrate time information into an attention model to embed medical concepts. Our work departs from Cai et al. ~\cite{MCE_Cai_2018} and Rajkomar et al.~\cite{Rajkomar_Google_2018} in that TeSAN integrates the time intervals between visits with expressive and multi-dimensional vectors into the context of medical concepts to capture the temporal relationships.

\section{Preliminary}\label{Preliminary}

This section begins by giving several definitions for medical concepts and targeted tasks. Because of the similarity between word embedding in the natural language processing literature and the code embedding of medical concept in EHRs, we then adopt some of the concepts and approaches designed for NLP tasks to apply to EHRs. We first introduce the concept of word embedding \cite{Mikolov_2013} to learn low-dimensional real-value distributed vector representations for medical concepts instead of discrete medical codes for downstream tasks; second, we adopt a sophisticated self-attention mechanism \cite{vaswani2017attention} for EHRs to capture the contextual information and temporal dependencies between the medical concepts for the context-aware medical concept representation, to achieve better empirical performance; lastly, the attention pooling \cite{liu2016learning} technique is leveraged to attentively select important elements from a set of input code embeddings, which is aimed at sequence compression or embedding via parameterized weighted sum. 

\subsection{Definitions}

\begin{definition}[Medical Concept] 
A medical concept is defined as a term or code to describe diagnosis, procedure, medication, and laboratory tests for an inpatient during a treatment process. We denote the set of medical concepts (e.g., ICD 585.5 for diagnosis, CPT 2001 for procedure) as $C$. 
\end{definition}

\begin{definition}[Visit]
A visit for an inpatient refers to a treatment process from admission to discharge, including an admission time stamp. We denote a visit as $V_{i,j} =<\textbf{\textit{x}}_{i,j},\textbf{\textit{t}}_{i,j}>$, where \textit{i} is the \textit{i}-th patient, \textit{j} the \textit{j}-th visit of the patient, $\textbf{\textit{x}}_{i,j} = [x^{i,j}_1,x^{i,j}_2,...,x^{i,j}_K] $, $\textbf{\textit{t}}_{i,j} = [t^{i,j}_1,t^{i,j}_2,...,t^{i,j}_K]$, \textit{K} is the number of medical concepts in a visit, $x^{i,j}_k$ is a medical concept and $t^{i,j}_k$ is admission time, where $k \in \{1,2, \dots, K\}$. 
\end{definition}

\begin{definition}[Patient Journey] 
A patient journey consists of a sequence of visits over time, which is denoted as $J_i = [V_{i,1},V_{i,2},...,V_{i,M}]$ where \textit{M} is the total number of visits for patient \textit{i}.
\end{definition}

\begin{definition}[Temporal Interval] \label{interval}
Temporal interval refers to difference in days between two visits in a patient journey, denoted as $\bigtriangleup_{jl} = |t^{i,j}_k - t^{i,l}_q|$, where $j,l \in \{1,...,M\}$ and $k,q \in \{1,....,K\}$.
\end{definition}

\begin{definition}[Problem] 
Given a set of patient journeys \textit{Js}, the problem is to learn an embedding function $f_C : C \xrightarrow{} R^d$ that maps every code in the set of  medical concept \textit{C} to a real-value dense vector with dimension \textit{d}.
\end{definition}

In this paper, a patient's medical data is stored to a sequence by chronologically concatenating \textit{M} visits in patient journey $J_i$. We will therefore ignore the indexes \textit{i,j} (which index the patients and their visiting times) for simplification, if it is possible to do so without causing confusion.

\subsection{Medical Concept Embedding}

Medical concept embedding is a fundamental processing unit in deep neural network-based EHRs. It transfers each discrete medical concept 
% (e.g., diagnosis code, treatment procedure code and drug code in pharmacy) 
into a distributed real-value vector representation. Formally, given a sequence or set of medical concepts $\textit{\textbf{x}}=[x_1, x_2, ..., x_n]\in \mathbb{R}^{|C| \times n}$, where $x_i$ is a one-hot vector, $|C|$ is the vocabulary size of the medical concept codes, and $n$ is the sequence length. A word embedding method (typically in the NLP literature, e.g. word2vec \cite{Mikolov_2013_b,Mikolov_2013}) is applied to the sequence, which outputs a sequence of low dimensional vectors $\textit{\textbf{c}} = [c_1, c_2, ..., c_n] \in \mathbb{R}^{d \times n}$, where $d$ is the embedding dimension of $c_i$. This process can be formally written as $\textbf{\textit{c}} = W^{(e)}\textit{\textbf{x}}$, where $W^{(e)} \in \mathbb{R}^{d \times |C|}$ is the embedding weight matrix that can be fine-tuned during the training phase. Following the idea of word embedding and context modeling in NLP, this paper is an attempt to embed medical concepts in low-dimensional vectors.  %  using temporal multi-dimensional attention via supervised-learning tasks.

%Word embedding is the basic processing unit in most DNN for sequence modeling. It transfers each discrete word into a representation vector of real values. Given a sequence of words (e.g., words or characters) $w=[w_1, w_2, ..., w_n]\in \mathbb{R}^{N \times n}$, where $w_i$ is a one-hot vector, \textit{N} is the vocabulary size and \textit{n} is the sequence length. A word embedding (e.g. word2vec \cite{Mikolov_2013_b,Mikolov_2013}) is applied to \textit{w}, which outputs a sequence of low dimensional vectors $x = [x_1, x_2, ..., x_n] \in \mathbb{R}^{d \times n}$, where \textit{d} is the feature size of $x_i$. This process can be formally written as $x = W^{(e)}w$, where $W^{(e)} \in \mathbb{R}^{d \times n}$ is the embedding weight matrix that can be fine-tuned during the training phase. Following the idea of word embedding, the paper is an attempt to embed medical concepts to low dimensional vectors using temporal multi-dimensional attention.

\subsection{Attention Mechanism}
\subsubsection{Vanilla Attention}

Given an input context of medical concepts $\textit{\textbf{c}} = [c_1, c_2, ..., c_n]$ composed of concept embeddings and a vector representation of a query $q \in \mathbb{R}^d$, vanilla attention~\cite{Bahdanau_2014} computes the alignment score between $q$ and each concept $c_i$ using a compatibility function $f(c_i, q)$. A softmax function then transforms the alignment scores $\alpha \in \mathbb{R}^n$ to a probability distribution $p(z|\textit{\textbf{c}}, q)$, where \textit{z} is an indicator of which concept is important to \textit{q}. A large $p(z = i|\textit{\textbf{c}}, q)$ means that $c_i$ contributes important information to $q$. This attention process can be formalized as
\begin{equation}
% \label{sg_eq}
\alpha = [f(c_i, q)]_{i=1}^n,
\end{equation}
\begin{equation}
% \label{sg_eq}
p(z|\textit{\textbf{c}}, q) = softmax(\alpha).
\end{equation}
The output $s$ is the weighted average of sampling a concept according to its importance, i.e.,
\begin{equation}
% \label{sg_eq}
s = \sum_{i=1}^n p(z = i|\textit{\textbf{c}}, q)\cdot c_i.
\end{equation}

Additive attention~\cite{Bahdanau_2014,shang2015neural} is commonly-used attention mechanism in which the compatibility function $f(\cdot)$ is parameterized by a multi-layer perceptron (MLP), i.e., 
\begin{equation}
\label{add_attn}
f(c_i, q) = w^T \sigma (W^{(1)}c_i+W^{(2)}q + b^{(1)})+b,
\end{equation}
where $W^{(1)} \in \mathbb{R}^{d\times d}$, $W^{(2)} \in \mathbb{R}^{d\times d}, w \in \mathbb{R}^d$ are learnable parameters, and $\sigma(\cdot)$ is an activation function. In contrast to additive attention, multiplicative attention~\cite{sukhbaatar2015end,rush2015neural} uses cosine similarity as the compatibility function for $f(x_i, q)$, i.e.,
\begin{equation}
\label{nulti_attn}
f(x_i, q) = \langle~W^{(1)}x_i,~W^{(2)}q~\rangle,
\end{equation}

In practice, although additive attention is expensive in time cost and memory consumption, it usually achieves better empirical performance for downstream tasks.

To improve the context modeling capability of the attention module, a \textbf{multi-dimensional (multi-dim) attention mechanism}~\cite{shen2018disan} that uses a feature-wise alignment score has recently been proposed. The alignment score from the attention compatibility function is computed for each feature; the score of a concept pair is a vector rather than a scalar, so the score might be large for some features but small for others to model more subtle context and dependency relationship. Formally, $P_{ki} \triangleq p(z_k = i|\textbf{\textit{c}}, q)$ denotes the attention probability of $i$-th element on $k$-th feature dimension, where the attention score is obtained from a multi-dim compatibility function by replacing the weight vector $w$ with a weight matrix in Eq.(\ref{add_attn}). For simplicity, we ignore the subscript $k$ if this does not cause confusion. The attention result can be written as $s=\sum^n_{i=1}P_{\cdot i}\odot c_{i}$. In the remaining paper, we use the multi-dim compatibility function by default for rich expressive power and better performance for downstream tasks. 

%Unlike vanilla attention, in multi-dimensional (multi-dim) attention~\cite{shen2018disan}, the alignment score is computed for each feature, i.e., the score of a concept pair is a vector rather than a scalar, so the score might be large for some features but small for others. *Multi-dim attention has \textit{d} indicators $z_1, \dots, z_d$ for \textit{d} features. Each indicator has a probability distribution that is generated by applying softmax to the \textit{n} alignment scores of the corresponding feature. Hence, for each feature \textit{k} in each medical concept \textit{i}, we have $P_{ki} \triangleq p(z_k = i|\textit{\textbf{c}}, q)$ where $P \in \mathbb{R}^{d\times n}$.
% A large $P_{ki}$ means that the feature \textit{k} in medical concept \textit{i} is important to \textit{q}. The output of multi-dim attention is written as
%\begin{equation}
%\label{multi_dim}
%s = \left[\sum^n_{i=i}P_{ki}\textbf{\textit{c}}_{ki}\right]^d_{k=1},
%\end{equation}
%For simplicity, we ignore the subscript \textit{k} where no confusion is caused. Then, Eq.~\ref{multi_dim} can be rewritten as an element-wise product, i.e., $s=\sum^n_{i=1}P_{\cdot i}\odot c_{i\cdot}$. Here, $P_{\cdot i}$ is computed by the additive attention in Eq.~\ref{add_attn} where $w^T$ is replaced with a weight matrix $W \in \mathbb{R}^{d\times d}$, which leads to a score vector for each medical concept pair.

\subsubsection{Self-attention mechanism}\label{c2csa}

The self-attention mechanism~\cite{hu2017reinforced,vaswani2017attention,shen2018disan,lizheng2018hierarchical} can produce context-aware representations by exploring the contextual relationships between two medical concepts $c_i$ and $c_j$ from the same context $\textbf{\textit{c}}$. It is naturally compatible with medical concept embedding because unlike  the commonly-used recurrent neural network, the self-attention mechanism is order-insensitive, making it suitable for all the medical concepts in a single patient visit. The query $q$ in the attention compatibility function (e.g., multi-dim compatibility function) is replaced by $c_j$ , i.e.,
\begin{equation}
\label{code2code}
f(c_i, c_j) = W^T \sigma (W^{(1)}c_i+W^{(2)}c_j + b^{(1)})+b.
\end{equation}
Similar to $P$ in multi-dim attention, each input medical concept $c_j$ is associated with a probability matrix $P_j$ such that $P^j_{ki} \triangleq  p(z_k = i|\mathbi{c}, c_j)$. The output representation for each $c_j$ is  
\begin{equation}
\label{s_code2code}
s_j=\sum^n_{i=1}P^j_{\cdot i}\odot c_{i}
\end{equation}
The final output of self-attention is $\textbf{\textit{s}} = [s_1, s_2, \dots, s_n]$, each of which is the medical-concept-context embedded representation for each medical concept. However, a fatal defect in previous self-attention mechanisms applied to NLP tasks is that they cannot model the relative time interval between the medical concepts from different patient visits, even if equipped with positional encoding \cite{vaswani2017attention}.

\subsubsection{Attention Pooling}
Attention pooling~\cite{lin2017structured,liu2016learning} explores the importance of each medical concept to the entire context given a specific task. This is used to compress a sequence of medical concept embeddings from a visit or a patient to a single context-aware vector sequence embedding for downstream classification or regression. In particular, $q$ is removed from the common compatibility function which is formally written as the following equation. 
\begin{equation}
\label{source2code}
f(c_i) = W^T \sigma (W^{(1)}c_i + b^{(1)})+b.
\end{equation}

The multi-dim attention probability matrix $P$ is defined as $P_{ki} \triangleq  p(z_k = i|\mathbi{c})$. The final output of the attention pooling, which is used as the sequence encoding,
has a similar form as the aforementioned attention mechanism, i.e., 
\begin{equation}
\label{s_source2code}
s=\sum^n_{i=1}P_{\cdot i}\odot c_i
\end{equation}

\begin{figure*}[!htb]
\centering
\includegraphics[width=0.7\textwidth]{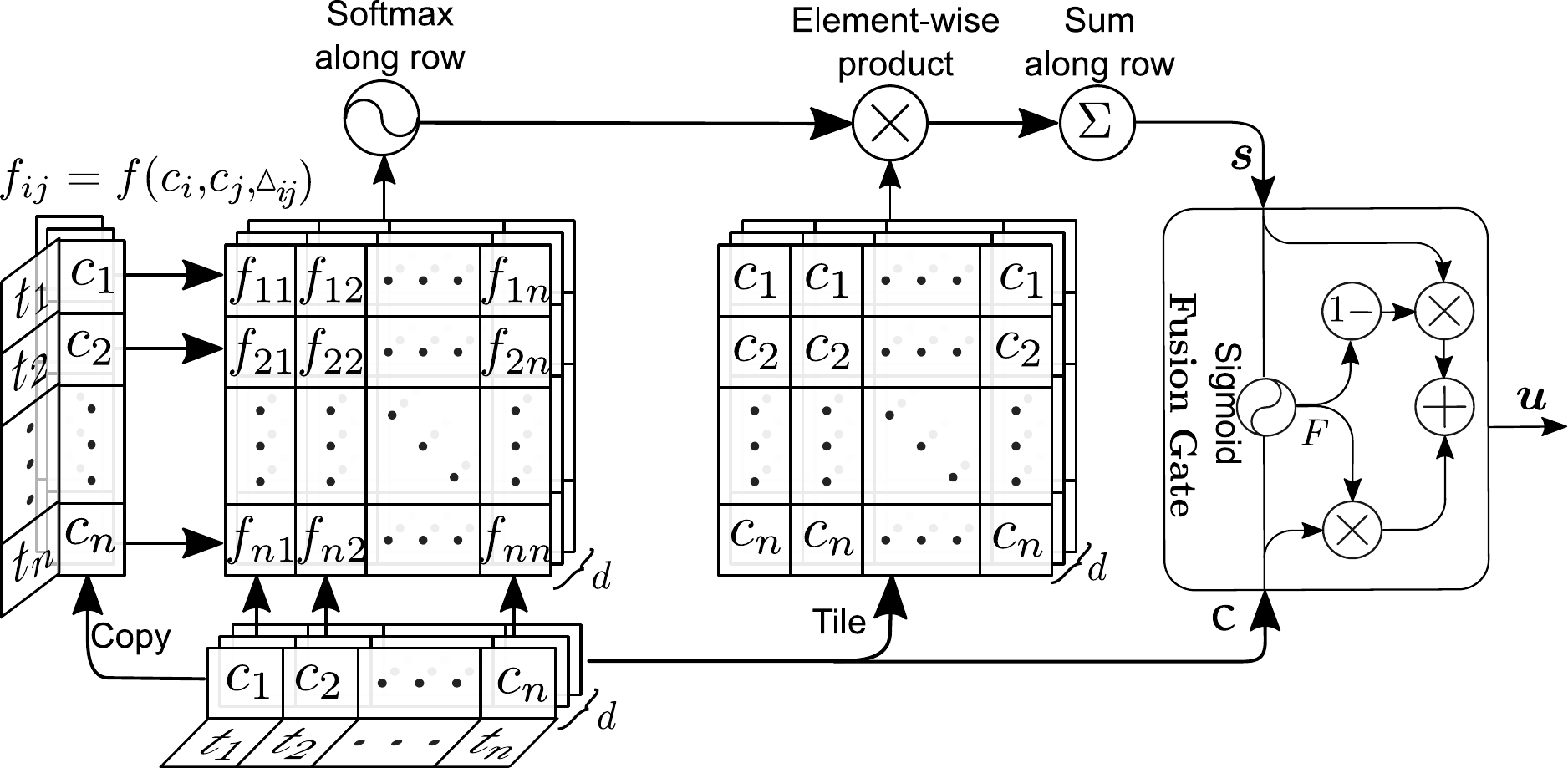}
\caption{Temporal self-attention mechanism. The inputs are embedded medical concepts ($c_1, c_2, \dots, c_n$) with corresponding visit time stamps $(t_1, t_2, \dots, t_n)$; $\bigtriangleup_{ij}$ can be obtained by giving $t_i$ and $t_j$. $f_{ij}$ is formally defined as Eq.(\ref{temp_code2code}); Softmax along row produces a probability distribution for each element of the multi-dim embedded medical concepts; Element-wise product outputs the weighted element-wise multi-dim medical concepts; Sum along row produces the weighted multi-dim medical concepts whose dimension size is the same as the size of the input of embedded medical concepts. The fusion gate merges the weighted output $\mathbi{s}$ and the input of embedded medical concepts $\mathbi{c}$ to produce output $\mathbi{u}$.} \label{temp_self_attn}
\end{figure*}

\section{Proposed Model}\label{Method}

We first introduce the ``temporal self-attention (TeSA)'' as a fundamental self-attention module. Then, we present the ``temporal self-attention network (TeSAN)'' for medical concept embedding, which uses TeSA as its context fusion module. Table~\ref{tab_notes} lists the notations used in the study.

\begin{table}[htbp]
\caption{Notations for HiSaNTH.}\label{tab_notes}
\centering
\begin{tabular}{|p{0.9cm}<{\centering}|p{6.8cm}|}
\hline
\textbf{Notation} & \textbf{Description} \\
\hline
$C$      & Set  of  unique medical concepts  \\ 
\hline
$|C|$     & The number of unique medical concepts \\ 
\hline
$V_{i,j}$        & The \textit{j}-th visit of the \textit{i}-th  patient\\ 
\hline
$\mathbi{x}_{i,j}$     & Set of medical concepts in $V_{i,j}$  \\ 
\hline
$x^{i,j}_k$   & The \textit{k}-th medical concept in $\mathbi{x}_{i,j}$, $k \in \{1,\dots, K\}$ \\ 
\hline
$\mathbi{c}_{i,j}$ & Set of medical concept embeddings in $\mathbi{x}_{i,j}$  \\ 
\hline
$c^{i,j}_k$   & The \textit{k}-th medical concept embedding in $\mathbi{c}_{i,j}$, $k \in \{1,\dots, K\}$ \\ 
\hline
$\mathbi{J}_{i}$       & A  patient  journey  consisting of  a  sequence  of  visits  over  time\\ 
\hline
% VSA  & 32.36& 58.16 & 14.42 & 45.47 \\ 
$\bigtriangleup$    & number of days between two visits in a patient journey \\
\hline
% VSA  & 32.36& 58.16 & 14.42 & 45.47 \\ 
$d$    & The embedding dimension \\
\hline
\textit{K}&  The number of medical concepts in a visit\\
\hline
\textit{M}&  The   total  number of visits  for  a patient\\
\hline
\end{tabular}
\end{table}

\subsection{Temporal Self-attention}
As discussed in the previous section, the self-attention mechanism is unlike the commonly-used   recurrent  neural   network which is order-insensitive and suitable  for  the  medical  concepts in  a  single  patient  visit. However, a flattened patient journey is a sequence of medical concepts with time stamps, so previous  self-attention  mechanisms  applied  to  NLP  tasks  cannot model the relative time interval between the medical concepts from different patient visits. Inspired by previous work on masked self-attention~\cite{shen2018disan,shen2018bi}, which achieves state-of-the-art performance on many NLP tasks, we propose a novel attention mechanism, called ``Temporal self-attention (TeSA)'', in which the attention mechanism captures the contextual information and temporal relationships between medical concepts.

Temporal self-attention is composed of a self-attention block to explore the contextual relationship and temporal interval and a fusion gate to combine the output and input of the attention block. Its structure is shown in Figure~\ref{temp_self_attn}. 
% Since self-attention is originally designed for NLP tasks, there is no consideration of temporal intervals within inputs, which are very important for modeling sequential patient visits to hospital. we propose a temporal interval-based self-attention, called ``temporal self-attention (TeSA)'', to capture the contextual relationship and temporal interval. 
We rewrite the self-attention in Eq.(\ref{code2code}) as a temporal-dependent format:
\begin{equation}
\label{temp_code2code}
f(c_i, c_j, \bigtriangleup_{ij}) = W^T \sigma (W^{(1)}c_i+W^{(2)}c_j + W^{(3)}e_{\bigtriangleup_{ij}}+ b^{(1)})+b
\end{equation}
where $\bigtriangleup_{ij}$ is the temporal days' interval between $t_i$ and $t_j$ as defined in Def.~(\ref{interval}), and $e_{\bigtriangleup_{ij}} \in \mathbb{R}^d$ is the temporal interval embedding, which is a learnable parameter. A temporal interval embedding layer is added before $e_{\bigtriangleup_{ij}}$ is taken as input to the TeSA module and the size of the embedding matrix is $\mathbb{R}^{n_{days}\times n}$, where $n_{days}$ is the number of days when the dataset spans.

Given input context \textit{\textbf{c}} and a temporal interval matrix $\bigtriangleup$, we compute $f(c_i, c_j, \bigtriangleup_{ij})$ according to Eq.(\ref{temp_code2code}), and follow the standard procedure of self-attention to compute the probability matrix $P_j$ for each $j \in [n]$. Each output $s_j$ in \textit{\textbf{s}} is computed as in Eq.(\ref{s_code2code}).

The final output $\textbf{\textit{u}} \in \mathbb{R}^{d\times n}$ of TeSA is obtained by combining the output \textit{\textbf{s}} and the input \textit{\textbf{c}} of the temporal self-attention block. This yields an encoded temporal interval and a context-aware vector representation for each medical concept. The combination is accomplished by a dimension-wise fusion gate, i.e.,
\begin{equation}
\label{gate}
F= sigmoid(W^{(f_1)}s+W^{(f_2)}c+b^{(f)})
\end{equation}
\begin{equation}
\label{fusion}
u = F\odot s + (1-F)\odot c
\end{equation}  
where $W^{(f1)},W^{(f2)} \in \mathbb{R}^{d\times d}$ and $b^{(f)} \in \mathbb{R}^d$ are the learnable parameters of the fusion gate.

\begin{figure}[!htb]
\centering
\includegraphics[width=0.22\textwidth]{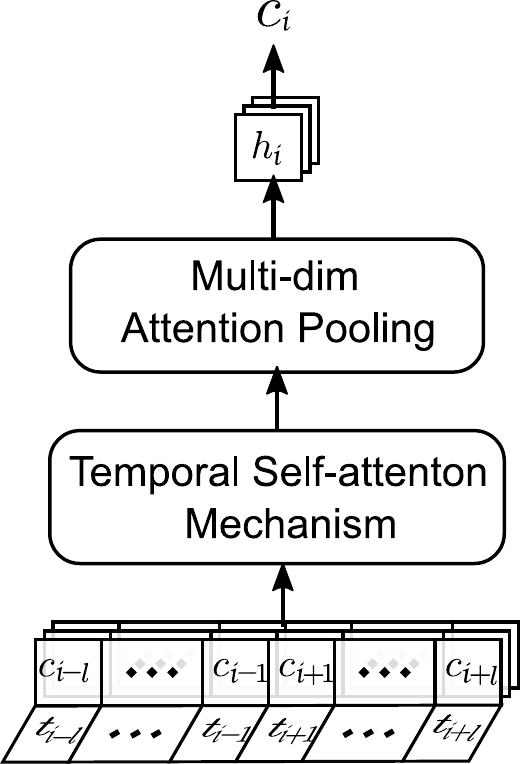}
\caption{Temporal self-attention network (TeSAN). Note that $c_i$ and $t_i$ are from concatenated visits in one patient journey, and $l$ is the size of the skip window.}  \label{tesan}
\end{figure}
\subsection{Temporal Self-attention Network}
We propose a light-weight network, ``Temporal Self-Attention Network (TeSAN)'' for  medical concept embedding. Its architecture is shown in Figure~\ref{tesan}. 

Given an input sequence of concept representation \textit{\textbf{c}}, which is from concatenated visits in one patient journey, TeSAN first applies the TeSA block to capture the contextual relationship and temporal interval information. The multi-dimensional attention pooling block takes the TeSA output as input to produce $h_i \in \mathbb{R}^d$ computed by Eq.~\ref{source2code} and \ref{s_source2code}. The context embedding result of TeSAN is exploited to predict target concept $c_i$.

\subsection{Loss Function}

The loss function is inspired by the  Word2Vec~\cite{Mikolov_2013,Mikolov_2013_b} model by using negative sampling to maximize

\begin{equation}
\label{NEG_eq}
J = \log \sigma ({c_i}^T {h_i})+\sum_{j=1}^r \mathbb{E}_{c_j \sim P(c)}[\log \sigma(-{c_j}^T {h_i})] ,
\end{equation}
where $\sigma$ is a Sigmoid function, $r$ is the number of negative samples, and $P(c)$ is the noise distribution~\cite{Mikolov_2013}.

\section{Experiments}\label{Experim}
The proposed model is evaluated on two public datasets via the unsupervised and prediction tasks. The source code of TeSAN is available at https://github.com/Xueping/tesan/.

\subsection{Dataset Description}
We conducted comparative studies on two public datasets listed as follows:

\begin{itemize}
\item \textbf{MIMIC III}~\cite{Johnson_2016} is an open-source, large-scale, de-identified EHRs data set consisting of clinical logs of patients admitted to intensive care units with serious conditions.
% . The MIMIC III~\cite{Johnson_2016} dataset mainly consists of clinical logs of patients admitted to critical care units with serious conditions.
The diagnosis codes in this dataset follow the ICD9 standard. The statistics of the dataset are provided in Tab.~\ref{tab1_sts}.

\item \textbf{CMS} is a publicly available\footnote{https://www.cms.gov} synthetic claims dataset, which includes four types of files: inpatient, outpatient, carrier and beneficiary summary. For our experiment, we chose only a subset of inpatient files between 2008 and 2010 as one of our two datasets. The basic statistical information is shown in Table~\ref{tab1_sts}.
\end{itemize}

\begin{table}[!htbp]
\caption{Statistics of Datasets.}\label{tab1_sts}
\centering
\scalebox{1}{
\begin{tabular}{|l|c|c|}
\hline
\textbf{Datasets}  & \textbf{MIMIC III} &  \textbf{CMS}\\
\hline
\# of patients & 46,520 & 755,214 \\
\hline
\# of visits  & 58,976 & 1,332,822\\
\hline
Avg. \# of visits per patient & 1.27 & 1.76 \\
\hline
\# of unique diagnosis codes & 6,985 & 7,873\\
\hline
\# of unique procedure codes & 2,032 & 10,726\\
\hline
\end{tabular}}

\end{table}

% \begin{table}[htbp]
% \caption{Table Type Styles}
% \begin{center}
% \begin{tabular}{|c|c|c|c|}
% \hline
% \textbf{Table}&\multicolumn{3}{|c|}{\textbf{Table Column Head}} \\
% \cline{2-4} 
% \textbf{Head} & \textbf{\textit{Table column subhead}}& \textbf{\textit{Subhead}}& \textbf{\textit{Subhead}} \\
% \hline
% copy& More table copy$^{\mathrm{a}}$& &  \\
% \hline
% \multicolumn{4}{l}{$^{\mathrm{a}}$Sample of a Table footnote.}
% \end{tabular}
% \label{tab1}
% \end{center}
% \end{table}

\subsection{Tasks of Clustering and Nearest Neighbour Search}
\subsubsection{Ground Truth}

Two clustering and nearest neighbour search (NNS)~\cite{MCE_Cai_2018} tasks were conducted to evaluate the quality of the medical concept embedding results. We selected the ground truth by using two well-organized ontologies, the ICD9 standard and Clinical Classifications Software (CCS)\footnote{https://www.hcup-us.ahrq.gov}. The ICD9 standard has a hierarchical structure~\cite{Wang_sen_2017} consisting of 19 categories. We used the high level nodes as the clustering labels. We obtained 19 categories for the MIMIC III and CMS datasets. Medical %For nearest neighbour search,%
concepts under the same subroot were considered as near neighbours for the nearest neighbour search. We obtained 555,873 near neighbour pairs for MIMIC III and 869,144 for CMS. This ground truth set is named \textbf{ICD}. CCS provides a way to classify diagnoses and procedures into a limited number of categories by aggregating individual ICD9 codes into broad diagnosis and procedure groups to facilitate statistical analysis and reporting\footnote{ https://www.hcup-us.ahrq.gov/toolssoftware/ccs/CCSUsersGuide.pdf}. CCS aggregates ICD9 diagnosis codes into 285 mutually exclusive categories. For clustering, we obtained 265 categories for MIMIC III and 267 for CMS. For the nearest neighbour search, we obtained 61,630 near neighbour pairs for MIMIC III and 89,546 for CMS. We refer to this ground truth set as \textbf{CCS}.

\subsubsection{Baseline Methods}

We compared our model with five baseline models that are state-of-the-art embedding methods as listed below. All baseline models were trained with their source codes. 
\begin{itemize}
\item \textbf{CBOW} \cite{Mikolov_2013_b} learns the representations by averaging the context within a sliding window to predict the target vector. 

\item \textbf{Skip-gram (Sg)} \cite{Mikolov_2013_b} predicts the target vector based on context, using each target word as an input to predict words within that context.

\item \textbf{GloVe}~\cite{Choi_2016} An unsupervised learning algorithm for obtaining vector representations for words. 

\item \textbf{med2vec}~\cite{Choi_2016} A multi-level embedding model for simultaneously embedding medical concepts and visits.

\item \textbf{MCE}~\cite{MCE_Cai_2018} A CBOW model with time-aware attention model to embed medical concepts with temporal information.

\item \textbf{TeSAN} Our proposed temporal self-attention network for medical concept embedding to capture the contextual relationship and temporal interval.
\end{itemize}
\subsubsection{Experimental Set-Up}
All infrequent medical concepts were removed and the threshold empirically set to 5. Patients whose number of hospital visits was less than 4 in CMS were empirically discarded. 
% The patients will be empirically discarded when the times of visiting to hospital is less than 4 in CMS. 
Following the original Word2vec~\cite{Mikolov_2013,Mikolov_2013_b}, the same negative sampling strategy as used in Skip-gram, CBOW
% VSA, 
and TeSAN, and the number of negative samples in MIMIC III and CMS was set to 10 and 5 respectively. All models were trained with 30 epochs for MIMIC III and 20 epochs for CMS. The dimension $d$ of the medical concept embedding was set to 100. The batch size is 64 for MIMIC III and 128 for CMS.

\subsubsection{Results}

We used the clustering and nearest neighbour search tasks to evaluate the embedding results on two public datasets: MIMIC III and CMS. We chose K-Means as the clustering algorithm, and used clustering performance indicator called Normalized Mutual Information (NMI), to evaluate the learned representations for the medical concepts. The skip window of our model was empirically set to 6 for MIMIC III and 7 for CMS. We used the two ground truth sets to evaluate the embedding performance of the proposed model and other baselines.

\begin{table}[!htb]
\caption{Clustering performance (NMI) of the models on two
datasets w.r.t. ground truths, ICD and CCS (\%).}\label{tab_tab2}
\centering
\scalebox{1}{
\begin{tabular}{|l|c|c|c|c|}\hline
\textbf{Model}&\multicolumn{2}{|c|}{\textbf{MIMIC III}} & \multicolumn{2}{|c|}{\textbf{CMS}} \\ 
\cline{2-5}
& \textbf{ICD} & \textbf{CCS} & \textbf{ICD} & \textbf{CCS}   \\
\hline
CBOW      & 25.34 & 53.09 & 08.52 & 41.70 \\ 
\hline
Sg        & 26.02 & 52.97 & 07.65 & 35.61\\ 
\hline
GloVe     & 17.68 & 46.57 & 06.50 & 33.45\\ 
\hline
med2vec   & 5.25 & 33.65 & 3.69 & 17.66\\ 
\hline
MCE       & 8.26 & 37.37 & 04.27 & 31.88\\ 
\hline
TeSAN    &\textbf{32.84} & \textbf{58.33} & \textbf{14.69} & \textbf{45.63}\\
\hline
\end{tabular}}
\end{table}

\begin{table}[!htb]
\caption{NNS performance (P@1) of the models on two
datasets w.r.t. ground truths, ICD and CCS (\%).}\label{tab_tab3}
\centering
\scalebox{1}{
\begin{tabular}{|l|c|c|c|c|}\hline
\textbf{Model}&\multicolumn{2}{|c|}{\textbf{MIMIC III}} & \multicolumn{2}{|c|}{\textbf{CMS}} \\ 
\cline{2-5}
& \textbf{ICD} & \textbf{CCS} & \textbf{ICD} & \textbf{CCS}   \\
\hline
CBOW      & 54.3 & 33.5 & 34.3 & 16.4 \\ 
\hline
Sg        & 52.0 & 35.2 & 29.1 & 10.1\\ 
\hline
GloVe     & 20.9 & 14.7 &  9.3 & 1.4\\ 
\hline
med2vec   & 11.8 & 4.8 & 10.7 & 2.8\\ 
\hline
MCE       & 11.8 & 3.0 &  24.7 & 1.5\\ 
\hline
TeSAN    &\textbf{66.1} & \textbf{43.8} & \textbf{47.8} & \textbf{24.9}\\
\hline            
\end{tabular}}
\end{table}

\begin{figure*}[!htb]
\centering
\includegraphics[width=1\textwidth]{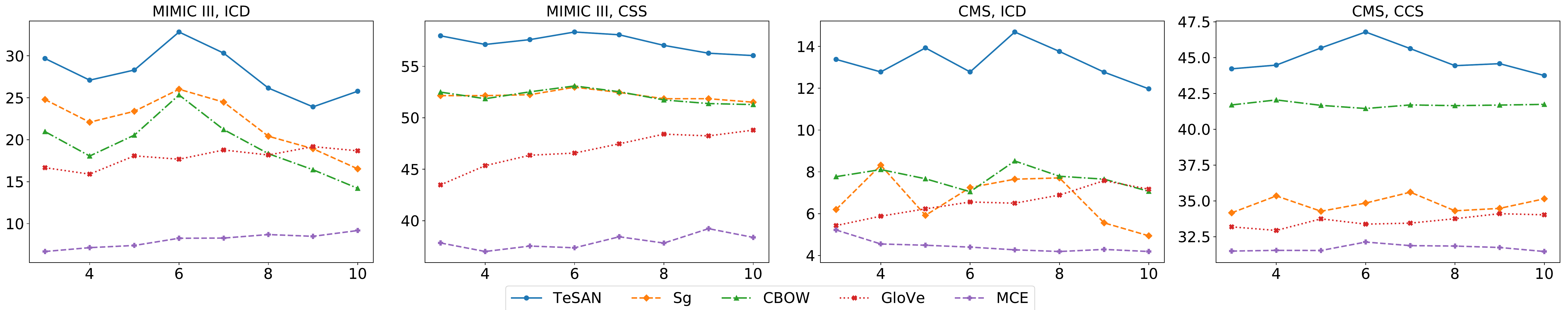}
\caption{NMI (\%) of clustering performance on two datasets w.r.t. two ground truths, ICD and CCS. The window size varies from 3 to 10.} \label{fig_mimic3_window}
\end{figure*}

\begin{figure*}[!htb]
\centering
\includegraphics[width=1\textwidth]{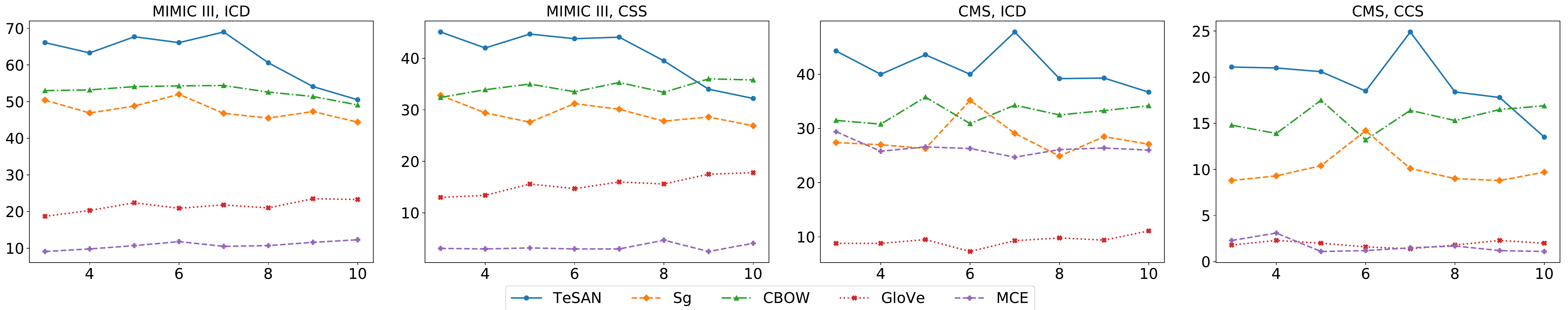}
\caption{P@1 (\%) of NNS performance on two datasets w.r.t. two ground truths, ICD and CCS. The window size varies from 3 to 10.} \label{fig_cms_window}
\end{figure*}

\paragraph{Overall Performance}
Normalized mutual information for clustering performance is reported in Table~\ref{tab_tab2}, and precision@1 (P@1) for NNS is shown in Table~\ref{tab_tab3}, where we highlight the best results. From the two tables, we find that the TeSAN model obtains the best performance in medical concept embedding compared to most state-of-the-art models on medical concept embedding. Our model outperformed the best baseline model for NMI by 6.82\% on ICD and 5.24\% on CCS over MIMIC III, and by 6.17\% on ICD and 3.93\% on CCS over CMS; for P@1 by 11.8\% on ICD and 8.6\% on CCS over MIMIC III, and by 13.5\% on ICD and 8.5\% on CCS over CMS. The superior performance of TeSAN over the other models can be explained by the introduction of the temporal self-attention model and the incorporation of the contextual information and temporal interval from the data, which creates a better learner of the medical concept embeddings. 

We find that the performance of med2vec and MCE are the worst in the clustering task and the NNS task in MIMIC III dataset, which indicates the importance of the skip window, since med2vec does not use the skip window and CME uses the skip window based on the weeks of the time unit. 

All models achieve better performance on the ground truth of CCS  than ICD for the clustering task, whereas the performance of all models on ICD is better than CSS for the NNS task. This might be explained by the fact that each of the well-organized ontologies have particular advantages for different tasks. We also find that the performance of all models on MIMIC III is better than those on CMS. There are two possible reasons: one is that the number of epochs is larger for MIMIC III than for CMS, and the other is that MIMIC III is drawn from real world healthcare data whereas CMS is synthesized data.

\paragraph{Performance of varying skip window sizes}  
To take the effects of the context window on the performance of the baseline and proposed models into consideration, we vary the size of the context window to compare performance. In this work, we only compare the proposed model TeSAN with other baselines; the exception is med2vec, due to lack of a parameter for window size. The window size is adjusted from 3 to 10. 

The results on the clustering on both datasets are summarized in Fig.~\ref{fig_mimic3_window}. The performance of most models is decreased, as an increase in window size induces noise. However, because GloVe makes use of global co-occurrences and MCE obtains a bigger skip window, neither is sensitive on increasing window size. As the window size is increased, GloVe and MCE achieve better performance with the larger window size. Moreover, the TeSAN model always outperforms the other models in terms of NMI on the MIMIC III dataset and in P@1 on CMS, which demonstrates that the integration of the proposed embedding model captures more comprehensive relationships between medical concepts. 

Figure~\ref{fig_cms_window} is the summary of results on the NNS task over two datasets. The TeSAN model outperforms the baseline models in terms of P@1 on the ground truth of CCS when the size of the skip window is not more than 8, which demonstrates that the attention mechanism benefits embedding in a smaller window. The performance of TeSAN slowly increases to the highest value and then quickly decreased, whereas the performance of GloVe and MCE is relatively stable over an increasing window size, which follow the same trend as their performance in the clustering task.  
%Other models obtain the local minimum values at skip window setting to 30, and achieve the best performances with window size setting to 6 for MIMIC III and 7 for CMS, respectively. 

\paragraph{Ablation Study.} 

We performed a detailed ablation study to examine the contributions of the proposed model components to unsupervised tasks. There are three replaceable components in this model: 
\begin{itemize}
\item  \textbf{Normal\_Sa:} we replaced the temporal self-attention module with a normal self-attention module;
\item  \textbf{Interval:} we only considered interval information in the temporal self-attention module; 
\item \textbf{Multi\_Sa:} we only considered contextual information in the temporal self-attention module; 
\item \textbf{TeSAN:}  is the proposed model. 
\end{itemize}

All models were trained with 30 epochs for MIMIC III and 20 epochs for CMS. The skip window of all models was empirically set to 6 for MIMIC III and 7 for CMS. Table~\ref{tab_tab4} and~\ref{tab_tab5} respectively show the performance for clustering and nearest neighbour search for the ablated models and our proposed model. 

From the two tables, we find that the TeSAN model obtains the best performance on medical concept embedding compared to the ablated models. Moreover, we note that Multi\_Sa outperforms Normal\_Sa, which gives us the confidence to apply multiple dimensional self-attention to learn the representation for medical concepts. It is clear that the Interval model provides comparable information with the learning embeddings of medical concepts to the performance of the Multi\_Sa and Normal\_Sa model. In particular, TeSAN outperforms the best ablated model for NMI by 1.22\% on ICD and 0.3\% on CCS over MIMIC III, and by 2.58\% on ICD and 1.88\% on CCS over CMS. It outperforms the best ablated model for P@1 by 1.3\% on ICD and 1.38\% on CCS over MIMIC III, and by 6.2\% on ICD and 6.3\% on CCS over CMS. 

\begin{table}[!htb]
\caption{Clustering  performance  (NMI)  of  the  models  on  two datasets w.r.t. ground truths, ICD and CCS (\%)}\label{tab_tab4}
\centering
\scalebox{1}{
\begin{tabular}{|l|c|c|c|c|}\hline
\textbf{Ablation}&\multicolumn{2}{|c|}{\textbf{MIMIC III}} & \multicolumn{2}{|c|}{\textbf{CMS}} \\ 
\cline{2-5}
& \textbf{ICD} & \textbf{CCS} & \textbf{ICD} & \textbf{CCS}   \\
\hline
Normal\_SA  &  30.66  &  55.99  & 11.99  &  43.13\\
\hline
Interval & 30.63 & 57.37 & 10.81 & 42.75 \\ 
\hline
Multi\_SA & 31.66& 58.03 & 12.11 & 43.75\\ 
\hline
TeSAN    &\textbf{32.84} & \textbf{58.33} & \textbf{14.69} & \textbf{45.63}\\
\hline            
\end{tabular}}
\end{table}

\begin{table}[!htb]
\caption{NNS performance (P@1) of the models on two datasets w.r.t. ground truths, ICD and CCS (\%)}\label{tab_tab5}
\centering
\scalebox{1}{
\begin{tabular}{|l|c|c|c|c|}\hline
\textbf{Ablation}&\multicolumn{2}{|c|}{\textbf{MIMIC III}} & \multicolumn{2}{|c|}{\textbf{CMS}} \\ 
\cline{2-5}
& \textbf{ICD} & \textbf{CCS} & \textbf{ICD} & \textbf{CCS}   \\ 
\hline
Normal\_SA  &  60.3 &  38.0 &  37.1 & 13.3\\
\hline
Interval & 64.8 &  42.0 &   37.0 & 16.3 \\
\hline
Multi\_SA & 63.7 &  40.5 &   41.6 & 18.6\\ 
\hline
TeSAN    &\textbf{66.1} & \textbf{43.8} & \textbf{47.8} & \textbf{24.9}\\
\hline            
\end{tabular}}
\end{table}

The ability of TeSAN to outperform the ablated models benefits from the introduction of the temporal self-attention model and the incorporation of  contextual information and the temporal interval from the data, which enables better embeddings of medical concepts to be learnt.

\subsection{Mortality Prediction Task}

We predicted impending inpatient death, defined as the latest discharge disposition of ``hospital expire''~\cite{kellett2012validation,tabak2013using,yamana2015procedure}. Note that there is no corresponding ``hospital expire'' flag in the CMS dataset, so the mortality prediction was only conducted on MIMIC III data.

\subsubsection{Baseline Methods}

First, we applied Gated Recurrent Units (GRU)~\cite{choi2018mime} with the following embedding strategies to map visit embedding sequence $v_1, \dots , v_M$ to a patient representation \textit{h}: 

\begin{itemize}
% \item \textbf{Raw:} A single visit $V_m$ is represented by a binary vector $\mathbi{x}_m \in \{0, 1\}^{|C|}$. Only the dimensions corresponding to the codes occurring in that visit is set to 1, and the rest are 0.
\item \textbf{CBOW+:} For the visit embedding, we simply mean the CBOW embeddings of the medical concepts within the visit.

\item \textbf{Skip-gram (Sg+):} We performed the same process as CBOW+ but used Skip-gram vectors instead of CBOW vectors.

\item \textbf{GloVe+:}~\cite{Choi_2016} The same process as CBOW+, but using GloVe vectors instead of CBOW vectors.

\item \textbf{med2vec:} We used Med2Vec~\cite{Choi_2016} to learn visit embedding where the dimension is the same as other embedding strategy.

\item \textbf{MCE+:}~\cite{MCE_Cai_2018} The same process as CBOW+, but using MCE vectors instead of CBOW vectors.

\item \textbf{TeSAN+:} The same process as CBOW+, but using the vectors of the proposed model.
\end{itemize}

We applied logistic regression to the patient representation \textit{h} to obtain a value between 0 (Survivor) and 1 (Death). All  models were trained end-to-end. We reported the Area under the Precision-Recall Curve (PR-AUC) and Area under the Receiver Operating Characteristic (ROC-AUC) in the experiment, as PR-AUC is considered a better measure for imbalanced data like ours~\cite{davis2006relationship,yamana2015procedure}. All models were trained with 50,00 steps; the batch size is 128 and the RNN cell type is GRU.

\subsubsection{Results}

Table~\ref{tab_prediction_performance} shows the test loss, PR-AUC and ROC-AUC of all models on dataset  MIMIC III. We find that TeSAN again consistently outperforms all baseline models. Achieving high specificity in mortality prediction is relatively easy as there are many more negative samples than positive ones. However, correctly identifying positive cases while ignoring negative ones requires a model differentiate between positive cases. This means attending to the details of patient records, such as the relationship between the diagnosis codes and temporal intervals. This is why TeSAN demonstrates a significant improvement in PR-AUC and ROC-AUC. Also, we note that GloVe shows very poor PR-AUC and ROC-AUC, and we observe that medical concepts with low frequencies are assigned near-zero vectors in this model, which might explain its poor performance.

\begin{table}[!htb]
\caption{Mortality prediction performance on MIMIC III dataset.}\label{tab_prediction_performance}
\centering
\scalebox{1}{
\begin{tabular}{|l|c|c|c|}\hline
\textbf{Model} & \textbf{test loss} & \textbf{test PR-AUC} & \textbf{test ROC-AUC}   \\ 
\hline
CBOW+  &  0.6765 &  0.5251 &   0.7784\\
\hline
Sg+ & 0.6764 &  0.5276 &   0.7785 \\
\hline
GloVe+ & 0.6834 &  0.4172 &  0.6548  \\ 
\hline
med2vec & 0.6772 &  0.5217 &  0.7690  \\ 
\hline
MCE+ & 0.6767 &  0.5204 &  0.7630  \\ 
\hline
TeSAN+    &\textbf{0.6736} & \textbf{0.5544} & \textbf{0.8064}\\
\hline            
\end{tabular}}
\end{table}

\subsubsection{Ablation Study}

We performed a similar ablation study with unsupervised learning tasks to examine the contributions of the proposed model components to the prediction task. 
\begin{itemize}
\item  \textbf{Normal\_Sa+:} For the visit embedding, we simply mean the Normal\_Sa  embeddings  of  the  medical  concepts  within  the visit.
\item \textbf{Interval+:} We perform the same process as Normal\_Sa+, but use Interval vectors instead of Normal\_Sa vectors.
\item \textbf{Multi\_Sa+:} We perform the same process as Normal\_Sa+, but use Multi\_Sa vectors instead of Normal\_Sa vectors. 
\item \textbf{TeSAN:}  We perform the same process as Normal\_Sa+, but use our proposed model vectors instead of Normal\_Sa vectors.
\end{itemize}

Table~\ref{tab_ablation_prediction} shows the mortality prediction performance for the ablated models and our proposed model. As can be seen from the table, the proposed model achieves the best performance compared to the ablated models on medical concept embedding. We observe the same trend as the ablated performance of unsupervised tasks, in which Multi\_Sa+ outperforms Normal\_Sa+. TeSAN outperforms the best ablated model by 2.75\% on PR-AUC and 2.54\% on ROC-AUC over MIMIC III. 

\begin{table}[!htb]
\caption{Mortality prediction performance on MIMIC III dataset.}\label{tab_ablation_prediction}
\centering
\scalebox{1}{
\begin{tabular}{|l|c|c|c|}\hline
\textbf{Ablation} & \textbf{test loss} & \textbf{test PR-AUC} & \textbf{test ROC-AUC}   \\ 
\hline
Normal\_SA+  &  0.6762 &  0.5233 &   0.7759\\
\hline
Interval+ & 0.6764 &  0.5191 &   0.7778 \\
\hline
Multi\_SA+ & 0.6759 &  0.5269 &  0.7808  \\ 
\hline
TeSAN+    &\textbf{0.6736} & \textbf{0.5544} & \textbf{0.8064}\\
\hline            
\end{tabular}}
\end{table}

\subsection{Visualization}

We present the visualized sample medical concepts with 3-dimension T-SNE results from the learned 100-dimension embedding vectors using our proposed medical embedding model. We selected three out of 19 categories in the high level of ICD9 standard and used the high level nodes as the clustering labels.

\begin{figure}[!htb]
\centering
\includegraphics[width=0.35\textwidth]{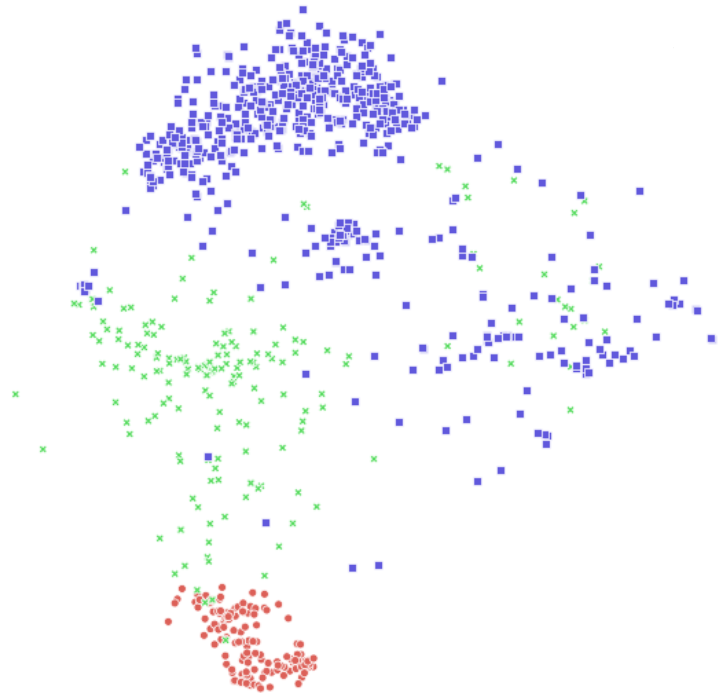}
\caption{Visualisation of 3 diagnosis categories in MIMIC III dataset.} \label{fig_mimic3_visual}
\end{figure}

Figure~\ref{fig_mimic3_visual} shows the 2D T-SNE results of three categories of sample medical concepts trained on MIMIC III, in which the red dots represent ``congenital anomalies'', the green dots represent ``certain conditions originating in the perinatal period'', and the blue dots represent ``symptoms, signs, and ill-defined conditions''. We find that the majority of the red and blue dots can be grouped in dense areas, whereas some green dots mix with the blue dots but are mainly clustered in a sparse area.

\begin{figure}[!htb]
\centering
\includegraphics[width=0.35\textwidth]{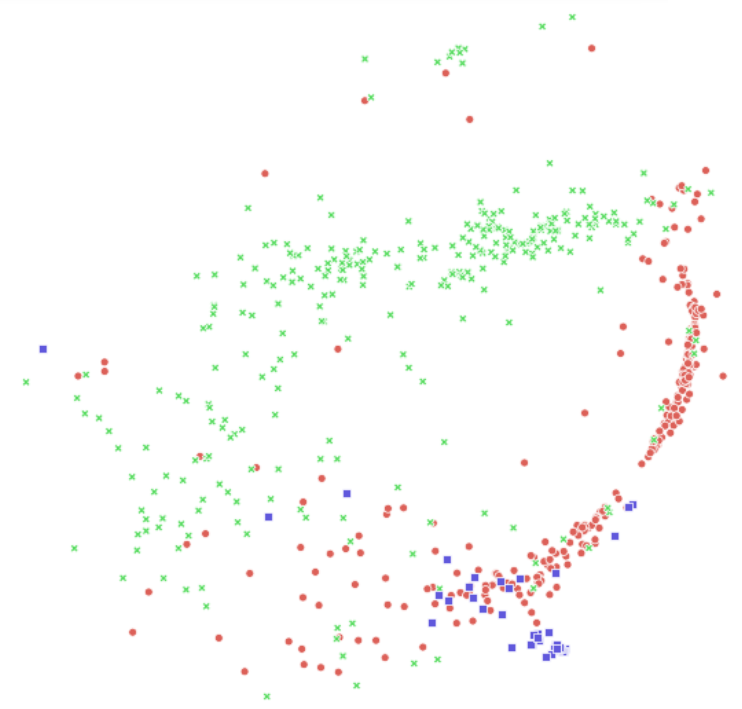}
\caption{Visualisation of 3 diagnosis categories in CMS dataset.} \label{fig_cms_visual}
\end{figure}

Figure~\ref{fig_cms_visual} shows the 2D T-SNE results three categories of sample medical concepts trained on CMS, in which the red dots represent ``diseases of the blood and blood-forming organs'', the green dots represent ``diseases of the respiratory system'', and the blue dots represent ``diseases of the genitourinary system''. We observe that although some green and red dots are intermingled, the red dots are grouped into a long dense arc area, and most green dots are grouped in another long dense area. The blue dots are clustered into a dense area close to red dots.

\section{Conclusion}\label{Conc}
This paper proposes a novel embedding model, the temporal self-attention network \textbf{TeSAN}. First, the model uses the self-attention mechanism to capture the contextual information and temporal relationships between medical concepts to compress the context into a vector representation by exploiting the temporal self-attention module (TeSA). Our model then applies the learning context vector to predict the target medical concept to learn representation for each medical concept. We conducted two types of tasks in unsupervised learning and prediction to evaluate the performance of our proposed model against baseline methods.We also executed ablation studies to examine the contributions of the proposed model components.  The experimental study demonstrates that the proposed model outperforms the baseline methods over two public datasets in tasks of clustering, nearest neighbour search, and mortality prediction.

\section{Acknowledgements}
This research was funded by the Australian Government through the Australian Research Council (ARC) under grants 1) LP160100630 partnership with Australia Government Department of Health and 2) LP150100671 partnership with Australia Research Alliance for Children and Youth (ARACY) and Global Business College Australia (GBCA). We also acknowledge the support of NVIDIA Corporation
and MakeMagic Australia with the donation of GPUs.

\bibliographystyle{IEEEtran}
\bibliography{References.bib}

\end{document}